\documentclass{article}

\usepackage{arxiv}

\usepackage{float} 
\usepackage[utf8]{inputenc} 
\usepackage[T1]{fontenc}    
\usepackage{hyperref}       
\usepackage{url}            
\usepackage{booktabs}       
\usepackage{amsfonts}       
\usepackage{nicefrac}       
\usepackage{microtype}      
\usepackage{lipsum}		
\usepackage{graphicx}
\usepackage{natbib}
\usepackage{doi}
\usepackage{tabularx} 
\usepackage{booktabs} 
\usepackage{algorithm}
\usepackage{algpseudocode}
\usepackage{amsmath}

\title{Kolmogorov-Arnold Network for Satellite Image Classification in Remote Sensing}

\renewcommand{\shorttitle}{\textit{arXiv} Template}

\author{
    Minjong Cheon \\
    \texttt{jmj2316@kist.re.kr} \\
}

\hypersetup{
pdftitle={A template for the arxiv style},
pdfsubject={q-bio.NC, q-bio.QM},
pdfauthor={Minjong},
pdfkeywords={First keyword, Second keyword, More},
}

\begin{document}
\maketitle

\begin{abstract}
In this research, we propose the first approach for integrating the Kolmogorov-Arnold Network (KAN) with various pre-trained Convolutional Neural Network (CNN) models for remote sensing (RS) scene classification tasks using the EuroSAT dataset. Our novel methodology, named KCN, aims to replace traditional Multi-Layer Perceptrons (MLPs) with KAN to enhance classification performance. We employed multiple CNN-based models, including VGG16, MobileNetV2, EfficientNet, ConvNeXt, ResNet101, and Vision Transformer (ViT), and evaluated their performance when paired with KAN. Our experiments demonstrated that KAN achieved high accuracy with fewer training epochs and parameters. Specifically, ConvNeXt paired with KAN showed the best performance, achieving 94\% accuracy in the first epoch, which increased to 96\% and remained consistent across subsequent epochs. The results indicated that KAN and MLP both achieved similar accuracy, with KAN performing slightly better in later epochs. By utilizing the EuroSAT dataset, we provided a robust testbed to investigate whether KAN is suitable for remote sensing classification tasks. Given that KAN is a novel algorithm, there is substantial capacity for further development and optimization, suggesting that KCN offers a promising alternative for efficient image analysis in the RS field.\end{abstract}


\section{Introduction}
In the field of remote sensing (RS), the rapid development of satellite technology has led to the proliferation of high-resolution RS images \citep{yang2018high}\citep{ma2015remote}. The integration of data from multiple satellites into time series has improved the detection of specific events and trends, enhancing our understanding of global climate patterns and ecological changes \citep{xu2019detecting}\citep{milesi2020measuring}. Satellites like Landsat, Sentinel, MODIS, and Gaofen are currently some of the most influential in Earth observation, with applications ranging from monitoring deforestation to tracking atmospheric conditions \citep{ustin2021current}. 
For example, the advent of sun-synchronous polar orbiting satellites has been critical for reliable and interpretable land surface observations, assisting in the monitoring of vegetation, crop yields, and forest disturbances \citep{ustin2021current}\citep{leblois2017has}. Moreover, these datasets have been extensively analyzed using deep learning models, which have demonstrated remarkable success in various applications such as land cover classification, object detection, and environmental monitoring \citep{shafique2022deep}\citep{adegun2023review}.  
Originally, the RS field was one of the first to actively apply deep learning techniques, paving the way for advanced image processing and analysis. Today, big firms such as Nvidia, Microsoft, Google, and Huawei are applying state-of-the-art deep learning models for climate prediction \citep{bodnar2024aurora} \citep{nguyen2023climax} \citep{nguyen2023scaling} \citep{pathak2022fourcastnet} \citep{lam2022graphcast} \citep{bi2023accurate}. This trend indicates that the RS field will also be significantly impacted by these novel deep learning algorithms. The integration of advanced deep learning techniques, including convolutional neural networks and transformers, is expected to further enhance the analysis and interpretation of remote sensing data. These advancements, already seen in weather forecasting, are likely to drive significant improvements in the accuracy and scope of remote sensing applications \citep{cheon2024karina}\citep{ham2019deep}\citep{ham2021unified}\citep{yuan2020deep}. 

However, applying deep learning to high-resolution datasets presents challenges. One of the primary issues is the enormous number of parameters required by deep neural networks, especially when dealing with high-resolution images \citep{lucas2018using}. However, applying deep learning to high-resolution datasets raises major obstacles. The vast number of parameters required for deep neural networks, particularly for high-resolution images, can result in higher processing costs and memory needs, rendering the training process inefficient and potentially infeasible with standard hardware resources. Furthermore, over-parameterization can result in more lengthy training times and potential overfitting, in which the model memorizes the training data rather than generalizes from it, limiting its effectiveness in real-world applications \citep{liu2021we}. To solve these issues, fresh approaches to effective training are required. A primary goal is to reduce the number of parameters while maintaining model performance. Model pruning, quantization, and using more compact network designs have all been investigated, but each has its tradeoffs and drawbacks \citep{hoefler2021sparsity}\citep{vadera2022methods}. Another possible avenue is to incorporate a novel network such as the Kolmogorov-Arnold Network (KAN), which could provide a more efficient learning framework \citep{liu2024kan}. In this paper, we explore the application of KAN architecture to RS scene classification tasks. By utilizing and comparing multiple pre-trained CNN and Vision Transformer (ViT) models, we aim to identify the most suitable pairings for the KAN framework. The key contributions of this article can be summarized as follows:

\begin{enumerate}
    \item This paper utilized the Kolmogorov-Arnold Network (KAN) architecture, replacing traditional MLPs for remote sensing (RS) scene classification tasks. 
    
    \item By utilizing and comparing multiple pre-trained CNN and Vision Transformer (ViT) models, we identified the most suitable pairings for the KAN. 
    
    \item Additionally, through comparisons involving various parameter configurations, both extensive and limited, we assessed the efficiency of the KAN, revealing its potential applicability across diverse fields.
    
\end{enumerate}

\section{Related Works}
Khan and Basalamah demonstrated the effectiveness of a multi-branch system combining pre-trained models for local and global feature extraction in satellite image classification, achieving superior accuracy across three datasets \citep{khan2023multi}. Chen et al. applied knowledge distillation to improve small neural networks' performance, enhancing accuracy on multiple remote sensing datasets \citep{chen2018training}. Broni-Bediako et al. used automated neural architecture search to develop efficient CNNs for scene classification such as EuroSAT, and BigEarthNet, outperforming traditional models with fewer parameters \citep{broni2021searching}. Temenos et al. introduced an interpretable deep learning framework using SHAP for land use classification, achieving high accuracy and enhanced interpretability \citep{temenos2023interpretable}. Yadav et al. improved classification on the EuroSAT dataset using pre-trained CNNs and advanced preprocessing techniques, with GoogleNet showing the best performance. Together, these studies highlight advancements in model accuracy, efficiency, and interpretability for remote sensing image classification \citep{yadav2024satellite}.

Vaca-Rubio et al. demonstrated that KANs outperformed conventional MLPs in satellite traffic forecasting, achieving higher accuracy with fewer parameters. An ablation study highlighted the significant impact of KAN-specific parameters, underscoring their potential for adaptive forecasting models \citep{vaca2024kolmogorov}. Bozorgasl and Chen developed Wavelet Kolmogorov-Arnold Networks (Wav-KAN) for federated learning, addressing heterogeneous data distribution among clients. Their integration of wavelet-based activation functions significantly enhanced computational efficiency, robustness, and accuracy, with extensive experiments on datasets such as MNIST, CIFAR10, and CelebA validating these improvements \citep{bozorgasl2024wav}. Abueidda and Pantidis created DeepONets using KANs to develop efficient surrogates for mechanics problems. Their approach, requiring fewer learnable parameters than traditional MLP-based models, was validated in computational solid mechanics, demonstrating notable computational speedups and efficiency, making it suitable for high-dimensional and complex engineering applications \citep{abueidda2024deepokan}.

While prior studies used pre-trained models, knowledge distillation, NAS, and interpretability frameworks to improve performance on remote sensing datasets, our strategy focuses on discovering the potential for applying KAN to the RS dataset. Our proposed approach is appropriate considering KAN's recent advancement, which opens up new possibilities for further optimization in remote sensing applications. Unlike other methods that depend on preprocessing techniques or sophisticated multi-branch frameworks, our integration of KAN with ConvNeXt demonstrates its ability to efficiently handle remote sensing datasets, providing a promising alternative for future research in this field.

\section{Materials and Methods}

\subsection{Dataset Description and Preprocessing}
The EuroSAT dataset, proposed by Helber et al. \citep{helber2019eurosat}, is utilized for land cover and land use classification tasks. It consists of 27,000 images collected from various cities across 34 European countries, covering ten distinct classes, including AnnualCrop, Forest, HerbaceousVegetation, Highway, Industrial, Pasture, PermanentCrop, Residential, River, and SeaLake. Each class includes 2000 to 3000 images, with detailed class descriptions and corresponding labels provided in Table 1. The images, acquired from the Sentinel-2A satellite, cover complex land cover scenes with high intra-class variance and measure 64 × 64 pixels with a spatial resolution of 10 meters per pixel.

The dataset has been split into three sections for experimentation: 18,900 photos for training, 4050 images for validation, and 4050 images for testing. Images are resized to 224 × 224 pixels and normalized using the mean and standard deviation values of ImageNet (mean = [0.485, 0.456, 0.406] and std = [0.229, 0.224, 0.225]) as part of the preparation of the EuroSAT dataset. The transformations for the given data consist of normalization, converting to tensors, random flips of the horizontal and vertical axes, and random resizing of the cropped images. We used pre-trained CNN models, such VGG16, VGG19, ResNet, and ConvNeXt, which require input data in this particular format, therefore this preprocessing is crucial \citep{corley2023revisiting}.

\begin{figure}[h!]
	\centering
	\includegraphics[width=0.8\textwidth]{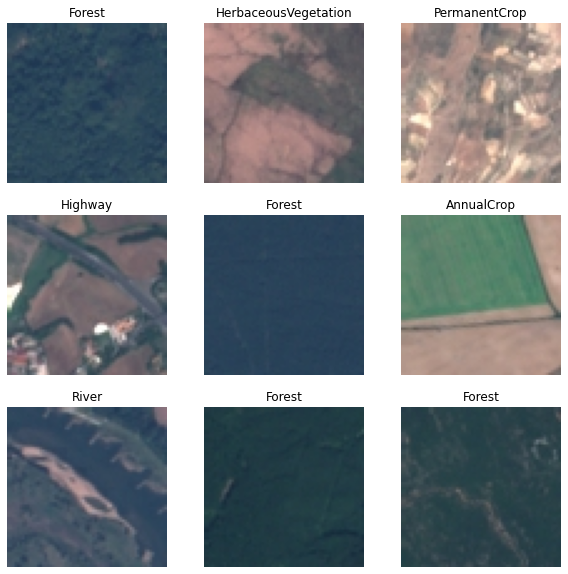}
	\caption{Example datasets from the EuroSAT used in the experiment}
	\label{fig:fig1}
\end{figure}

\subsection{ConvNeXt}
The ConvNeXt algorithm builds on standard ConvNet architectures, specifically modernizing the ResNet model by incorporating several design elements from Vision Transformers (VIT) \citep{liu2022convnet}. Key structural adjustments include modifying the stem cell to a patchify layer, adopting depthwise convolutions with increased network width, using inverted bottleneck blocks, and introducing large kernel sizes for convolutions. The patchify layer splits the input image into smaller, non-overlapping patches, treating each patch as a sequence token for further processing, similar to how words are treated in natural language processing \citep{dosovitskiy2020image}. Depthwise convolution processes each input channel separately, significantly reducing computational complexity, while pointwise convolution (1x1 convolution) combines information across channels \citep{howard2017mobilenets}. Inverted bottleneck blocks, popularized by MobileNetV2, expand the number of channels before reducing them back down, effectively capturing spatial and channel-wise features with reduced computational cost \citep{sandler2018mobilenetv2}. Additionally, ConvNeXt implements GELU activation functions, fewer normalization layers (switching from BatchNorm to LayerNorm), and reduces the number of activation functions per block. 

\subsection{Kolmogorov-Arnold Networks}
Kolmogorov-Arnold Networks (KANs), inspired by the Kolmogorov-Arnold representation theorem, is an advanced type of neural network featuring learnable activation functions on edges, unlike the fixed activations on nodes in traditional Multi-Layer Perceptrons (MLPs). These activation functions are parameterized by B-splines, which are piecewise polynomial functions defined by control points and knots. Each input feature \( x_p \) is transformed by spline-parameterized functions \( \varphi_{q,p} \), aggregated into intermediate values for each \( q \), and then passed through functions \( \Phi_q \). The final output \( f(\mathbf{x}) \) is the sum of these transformed values, allowing the network to flexibly and efficiently capture intricate data patterns. The activation functions in KANs are a combination of a Basis Function and a Spline, with the Basis Function often being the Sigmoid Linear Unit (SiLU), defined as \( \text{silu}(x) = \frac{x}{1 + e^{-x}} \). The spline component \( \text{spline}(x) = \sum_i c_i B_i(x) \) uses B-spline basis functions \( B_i(x) \) and coefficients \( c_i \), which are learned during training. These coefficients determine the final shape of the activation functions, replacing the need for traditional linear transformation parameters \( W \) and \( b \) in MLPs.

\begin{figure}[h!]
	\centering
	\includegraphics[width=0.8\textwidth]{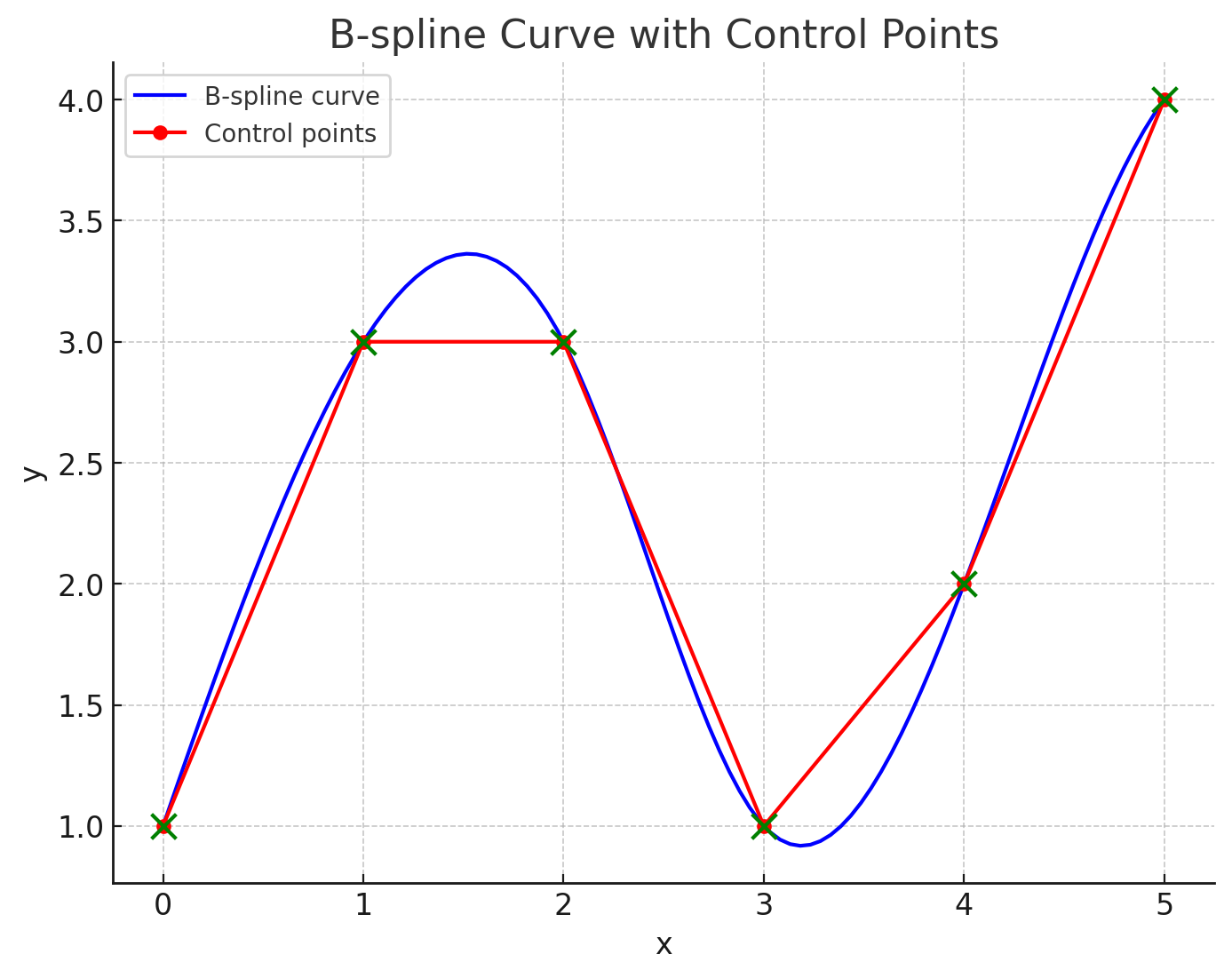}
	\caption{Visualization of a B-spline curve with its corresponding control points. The B-spline curve (blue) is smoothly fitted through a series of control points (red), which are marked with green crosses.}
	\label{fig:fig1}
\end{figure}

The formula for the Kolmogorov-Arnold Network (KAN) is given by:

\begin{equation}
f(\mathbf{x}) = \sum_{q=1}^{2n+1} \Phi_q \left( \sum_{p=1}^{n} \varphi_{q,p}(x_p) \right)
\end{equation}
\begin{center}
\text{The function \( f(\mathbf{x}) \) in a KAN, where \( \varphi_{q,p}(x_p) \) are spline functions and \( \Phi_q \) are transformations.}
\end{center}

\begin{equation}
\varphi(x) = w \left( b(x) + \text{spline}(x) \right)
\end{equation}
\begin{center}
\text{$\varphi(x)$ denotes the activation function, $w$ is a weight, $b(x)$ is the basis function and $\text{spline}(x)$ is the spline function.}
\end{center}

\begin{equation}
b(x) = \text{silu}(x) = \frac{x}{1 + e^{-x}}
\end{equation}
\begin{center}
\text{$b(x)$ is the basis function, implemented as $\text{silu}(x)$ (Sigmoid Linear Unit).}
\end{center}

\begin{equation}
\text{spline}(x) = \sum_i c_i B_i(x)
\end{equation}
\begin{center}
\text{$\text{spline}(x)$ is the spline function, $c_i$ are the coefficients, and $B_i(x)$ are the B-spline basis functions.}
\end{center}

\begin{table}[h!]
\centering
\caption{Comparison of Multi-Layer Perceptron (MLP) and Kolmogorov-Arnold Network (KAN)}
\begin{tabular}{|c|c|c|}
\hline
\textbf{Model} & \textbf{Multi-Layer Perceptron (MLP)} & \textbf{Kolmogorov-Arnold Network (KAN)} \\ \hline
\textbf{Formula (Deep)} & MLP($\mathbf{x}$) = $(\mathbf{W}_3 \circ \sigma_2 \circ \mathbf{W}_2 \circ \sigma_1 \circ \mathbf{W}_1)(\mathbf{x})$ & KAN($\mathbf{x}$) = $(\Phi_3 \circ \Phi_2 \circ \Phi_1)(\mathbf{x})$ \\ \hline
\end{tabular}
\end{table}

KANs yield various advantages over MLPs, including better accuracy and interpretability with fewer parameters. They achieve this by having smaller architectures that can perform comparably or better than larger MLPs in tasks such as data fitting and partial differential equation (PDE) solving.  Since KANs could be depicted, they are additionally helpful in discovering mathematical and physical laws in scientific applications. Moreover, KANs can aid in preventing catastrophic forgetting, a neural network problem when the learning of new information causes the loss of previously learned information. One notable drawback of KANs over MLPs is their slower training pace, which optimization can address \citep{liu2024kan}.

\subsection{Kolmogorov-Arnold Networks: KCN}
The proposed algorithm integrates the ConvNeXt architecture with KAN to enhance the learning capabilities of a pre-trained ConvNeXt model. The ConvNeXt model is first pre-trained and its layers are frozen to preserve the learned features. In place of the traditional MLP classifier, the model introduces two KANLinear layers. The KANLinear layers utilize learnable activation functions on the edges instead of fixed activation functions on the nodes, as in traditional neural networks. These activation functions are represented as B-splines, which offer more flexibility and stability. The KANLinear layer replaces linear weights with these spline functions, allowing the model to adapt more effectively to the input data. 
We used multiple strategies to evaluate the KAN's performance for RS classification tasks in remote sensing. First, we combined the KAN with different pre-trained CNNs and ViTs to find the best pairing. Secondly, in order to show the effectiveness of the KAN, we conducted tests in which we decreased the KAN's parameter count and contrasted the results with the original models. Third, we compared the performance evolution of the KAN model to that of the MLP by analyzing their accuracies per epoch. In this comparison, we demonstrated the potential of replacing the MLP with the KAN model for remote sensing datasets.

\begin{algorithm}[H]
\caption{\textbf{KCN: Integrating ConvNeXt with Kolmogorov-Arnold Networks}}
\begin{algorithmic}[]
\State \textbf{Input:} Pre-trained ConvNeXt model
\State \textbf{Output:} Trained ConvNeXtKAN model

\Procedure{KCN Training}{}
\State Load the pre-trained ConvNeXt model.
\State Freeze the layers of the ConvNeXt model.
\State Remove the original classifier from the ConvNeXt model.
\State Extract the number of features from the last layer of ConvNeXt.

\State Initialize \textit{kan1} with input size equal to the extracted features and output size of 32.
\State Initialize the second KANLinear layer (\textit{kan2}) with input size 32 and output size 10.

\State Define the \textit{forward pass}:
\State \quad Pass the input through the ConvNeXt model.
\State \quad Flatten the output from the ConvNeXt model.
\State \quad Pass the flattened output through \textit{kan1}.
\State \quad Pass the output from \textit{kan1} through \textit{kan2}.
\State \quad Return the final output.

\State Train the model using backpropagation and gradient descent.
\EndProcedure
\end{algorithmic}
\end{algorithm}

\section{Result}
First, we applied various CNN-based models that were pre-trained on the ImageNet dataset. Typically, these pre-trained networks utilize MLP layers for classification or regression tasks. However, since our research aimed to demonstrate the potential of replacing MLP with KAN, we substituted the MLP with KAN in these models. We evaluated several models, including VGG16, MobileNetV2, EfficientNet, ConvNeXt, ResNet101, and ViT. The observed accuracies for these models were 88\%, 75\%, 67\%, 94\%, 75\%, and 92\%, respectively, as shown in Figure 3. By comparing these networks, we discovered that ConvNeXt is the most suitable algorithm with KAN. Figure 4 below shows training loss per iteration at epoch 0. 

\begin{figure}[H]
\centering
\includegraphics[width=0.8\textwidth]{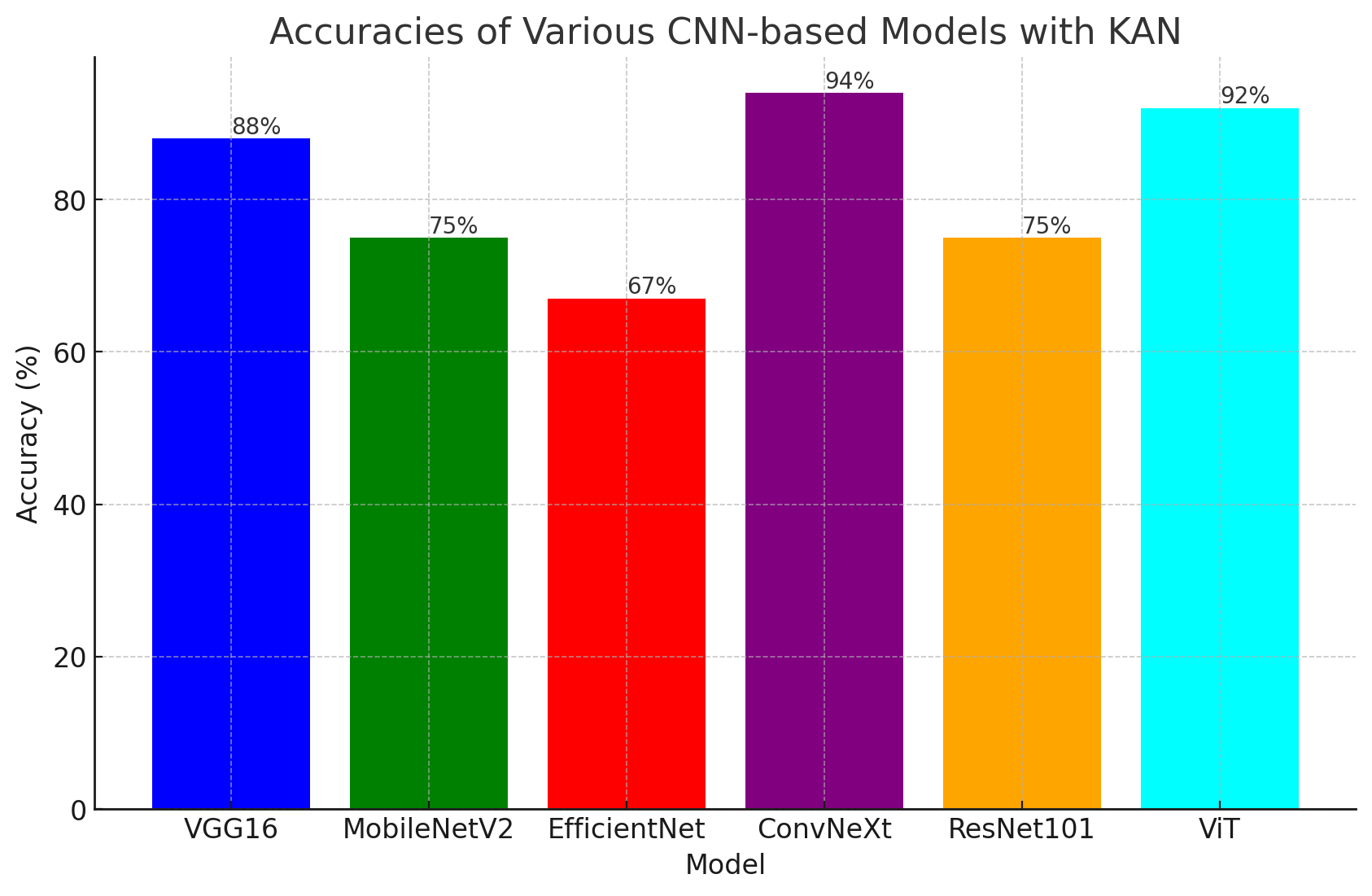}
\caption{Accuracies of Various CNN-based Models with KAN}
\label{fig:accuracy_per_epoch}
\end{figure}

\begin{figure}[H]
	\centering
	\includegraphics[width=0.8\textwidth]{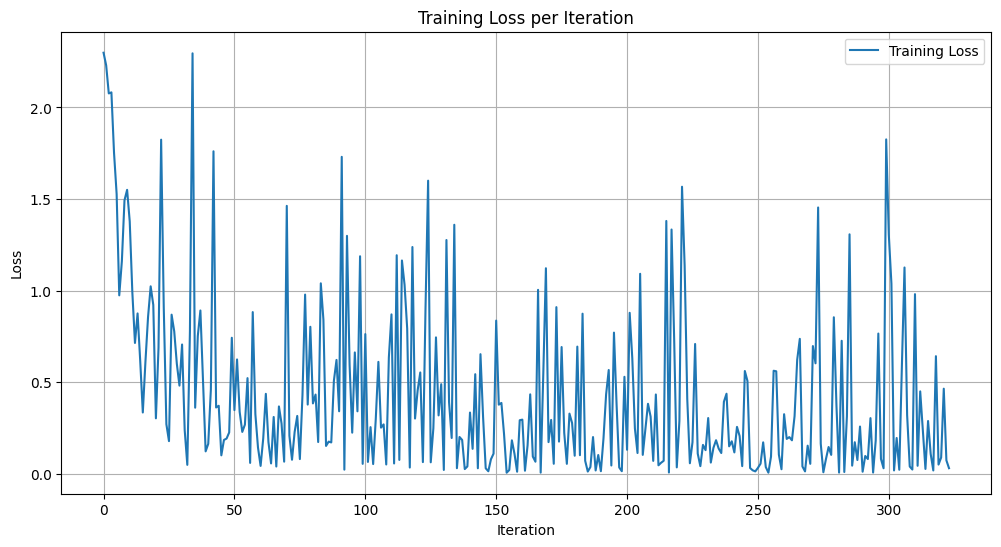}
	\caption{KCN's training loss per iteration at epoch 0.}
	\label{fig:fig3}
\end{figure}

To demonstrate the efficiency of KAN, we initially configured each KAN layer with 256 nodes and subsequently reduced the nodes to 32 for comparison. The results revealed identical performance between both configurations. Both setups achieved an accuracy of 94\% in the first epoch, which increased to 96\% in the second epoch, maintaining this accuracy in the subsequent epochs, as described in Table 2. This experiment indicates that KAN layers can achieve high accuracy with fewer training epochs, even when the number of nodes is significantly reduced and this proved the efficiency of the KAN layer.

\begin{table}[H]
\centering
\caption{Comparison of Accuracy per Epoch for KAN Layers with Different Node Configurations}
\begin{tabular}{@{}cccccc@{}}
\toprule
\textbf{Nodes} & \textbf{Epoch 1} & \textbf{Epoch 2} & \textbf{Epoch 3} & \textbf{Epoch 4} & \textbf{Epoch 5} \\ \midrule
\textbf{256} & 94\% & 96\% & 96\% & 96\% & 96\% \\
\textbf{32} & 94\% & 96\% & 96\% & 96\% & 96\% \\ \bottomrule
\end{tabular}
\end{table}

We also compared the classification accuracy per epoch using KAN and MLP architectures integrated with the ConvNeXt model. Figure 5 yielded that both KAN and MLP started with 94\% accuracy at the first epoch. KAN rapidly improved to 96\% by the second epoch and maintained this accuracy through the fifth epoch, while MLP reached 95\% by the second epoch and gradually achieved 96\% by the third epoch, sustaining it afterwards. This revealed that KAN attained high accuracy slightly more quickly than MLP, but both ultimately reached similar accuracy, suggesting there is not a significant difference between the two for remote sensing classification tasks.

\begin{figure}[H]
\centering
\includegraphics[width=0.8\textwidth]{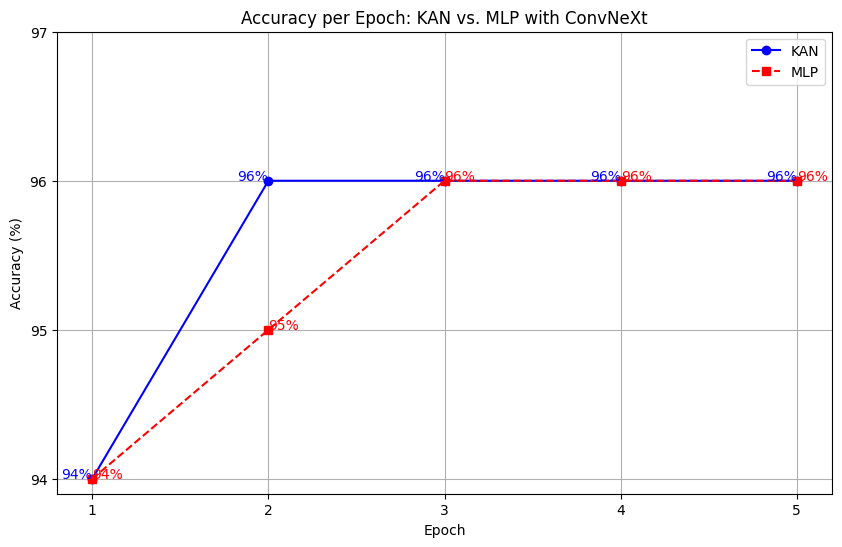}
\caption{Accuracy per Epoch: KAN vs. MLP with ConvNeXt}
\label{fig:accuracy_per_epoch}
\end{figure}

\section{Conclusion}
In this study, we introduced the integration of KAN with multiple pre-trained CNN-based models, such as VGG16, MobileNetV2, EfficientNet, ConvNeXt, ResNet101, and VIT, to investigate the best pairings for remote sensing (RS) scene categorization tasks. Our approach, KCN, revealed KAN's ability to outperform classic MLPs by obtaining excellent accuracy with fewer training epochs and parameters. The novel use of KAN in RS datasets is a big step forward, demonstrating the model's ability to efficiently handle RS data.

Our findings showed that KAN could replace standard MLPs, achieving satisfactory accuracy. This proves that KCN can successfully combine the strengths of KAN and CNNs, leading to desirable performance for RS scene categorization. Despite these results, our study acknowledges several shortcomings. One major limitation is the lack of evidence about the interpretability of KAN layers, which is crucial for understanding the decision-making process of the model. Additionally, since KAN is a relatively recent development, there is a need for further research to optimize its performance and integration into diverse remote sensing applications. Future research should address these issues by improving KAN's interpretability and investigating more effective integration solutions for various remote sensing applications.

To conclude, we utilized the KAN architecture to replace traditional MLPs for RS scene classification tasks. By utilizing and comparing multiple pre-trained CNN and ViT models, we identified the most suitable pairings for the KAN. Furthermore, through comparisons involving various parameter configurations, both extensive and limited, we assessed the efficiency of KAN, revealing its potential applicability across diverse fields instead of MLP. Our findings demonstrate the effectiveness of the KCN approach and its potential to influence the RS field by utilizing the advanced capabilities of KAN and CNN models.

\bibliographystyle{unsrtnat}
\bibliography{references}
\end{document}